\documentclass{llncs}
\usepackage{graphicx}
\begin{document}

{\let\thefootnote\relax\footnotetext{Copyright \textcopyright\ 2019 for this paper by its authors. Use permitted under Creative Commons License Attribution 4.0 International (CC BY 4.0). IberLEF 2019, 24 September 2019, Bilbao, Spain. }}

\title{Applying a Pre-trained Language Model to Spanish Twitter Humor Prediction}
\author{Bobak Farzin \inst{1} Piotr Czapla \inst{2} and Jeremy Howard\inst{3} \\June 2019}
\authorrunning{Farzin et al. 2019}
%
\institute{USF Data Institue, WAMRI Visting Scholar, USA \\
		\email{bfarzin@gmail.com} \\
		\and
		n-waves, Poland\\
		\email{Piotr.Czapla@n-waves.com}
\and  
University of San Francisco \& Fast.ai, USA\\ \email{j@fast.ai}}


\maketitle

\begin{abstract}
Our entry into the \textit{HAHA 2019 Challenge} placed $3^{rd}$ in the classification task and $2^{nd}$ in the regression task.  We describe our system and innovations, as well as comparing our results to a Naive Bayes baseline.
A large Twitter based corpus allowed us to train a language model from scratch focused on Spanish and transfer that knowledge to our competition model.  To overcome the inherent errors in some labels we reduce our class confidence with label smoothing in the loss function.
All the code for our project is included in a GitHub\footnotemark[4] repository for easy reference and to enable replication by others.

\end{abstract}

\section{Introduction}
\label{intro}
\newcommand{\chapquote}[3]{\begin{quotation} \textit{#1} \end{quotation} \begin{flushright} - #2, \textit{#3}\end{flushright} }

\begin{quote}
- !`Socorro, me ha picado una víbora!\\
- ?`Cobra?\\
- No, gratis.\footnotemark[5]
\end{quote}
Google Translation:
\begin{quote}
- Help, I was bitten by a snake!\\
- Does it charge?\\
- Not free.
\end{quote}
\pagebreak
\footnotetext[4]{https://github.com/bfarzin/haha\_2019\_final, Accessed on 19 June 2019}
\footnotetext[5]{https://www.fluentin3months.com/spanish-jokes/, Accessed on 19 June 2019}

Humor does not translate well because it often relies on double-meaning or a subtle play on word choice, pronunciation, or context.  These issues are further exacerbated in areas where space is a premium (as frequent on social media platforms), often leading to usage and development of shorthand, in-jokes, and self-reference. Thus, building a system to classify the humor of tweets is a difficult task.  However, with transfer-learning and the Fast.ai library\footnote[6]{https://docs.fast.ai/, Accessed on 19 June 2019}, we can build a high quality classifier in a foreign language. Our system outperforms a Naive Bayes Support Vector Machine (NBSVM) baseline, which is frequently considered a "strong baseline" for many Natural Language Processing (NLP) related tasks (see 
Wang et al \cite{Wang:2012:BBS:2390665.2390688}).

Rather than hand-crafted language features, we have taken an "end to end" approach building from the raw text to a final model that achieves the tasks as presented.  Our paper lays out the details of the system and our code can be found in a GitHub repository for use by other researchers to extend the state of the art in sentiment analysis. 

\paragraph{Contribution} Our contributions are three fold.  First, we apply transfer-learning of a language model based on a larger corpus of tweets.  Second, we use a label smoothed loss, which provides regularization and allows full training of the final model without gradual unfreezing.  Third, we select the best model for each task based on cross-validation and 20 random-seed initialization in the final network training step.

\section{Task and Dataset Description}
\label{sec:task}
The \textit{Humor Analysis based on Human Annotation (HAHA) 2019}\cite{overview_haha2019} competition asked for analysis of two tasks in the Spanish language based on a corpus of publicly collected data described in Castro et al.~\cite{castro2018crowd}:
\begin{itemize}
\item \textbf{Task1: Humor Detection}:Determine if a tweet is humorous. System ranking is based on F1 score which balances precision and accuracy.
\item \textbf{Task2: Funniness Score}:If humorous, what is the average humor rating of the tweet? System ranking is based on root mean-squared error (RMSE).
\end{itemize}
The HAHA dataset includes labeled data for 24,000 tweets and a test set of 6,000 tweets (80\%/20\% train/test split.)  Each record includes the raw tweet text (including accents and emoticons), a binary humor label, the number of votes for each of five star ratings and a ``Funniness Score'' that is the average of the 1 to 5 star votes cast.  Examples and data can be found on the CodaLab competition webpage\footnote[7]{http://competitions.codalab.org/competitions/22194/ Accessed on 19 June 2019}.

\section{System Description}
\label{sec:system}
We modify the method of Universal Langage Model Fine-tuning for Text Classification (ULMFiT) presented in Howard and Ruder~\cite{HowardRuder:DBLP:journals/corr/abs-1801-06146}.  The primary steps are: 
\begin{enumerate}
	\item Train a language model (LM) on a large corpus of data
	\item Fine-tune the LM based on the target task language data
	\item Replace the final layer of the LM with a softmax or linear output layer and then fine-tune on the particular task at hand (classification or regression)
\end{enumerate}
Below we will give more detail on each step and the parameters used to generate our system.
\subsection{Data, Cleanup \& Tokenization}
\label{sec:datacleaning}
\subsection{Additional Data}
We collected a corpus for our LM based on Spanish Twitter using tweepy\footnote[8]{http://github.com/tweepy/tweepy, Accessed on 19 June 2019} run for three 4-hour sessions and collecting any tweet with any of the terms 'el','su','lo','y' or 'en'. We excluded retweets to minimize repeated examples in our language model training.  In total, we collected 475,143 tweets - a data set is nearly 16 times larger than the text provided by the competition alone.  The frequency of terms, punctuation and vocabulary used on Twitter can be quite different from the standard Wikipedia corpus that is often used to train an LM from scratch.  

In the fine-tuning step, we combined the train and test text data \textit{without labels} from the contest data.
\subsection{Cleaning}
We applied a list of default cleanup functions in sequence (see list below).  They are close to the standard clean-up included in the Fast.ai library with the addition of one function for the Twitter dataset. Cleanup of data is key to expressing information in a compact way so that the LM can use the relevant data when trying to predict the next word in a sequence.
\begin{enumerate}
	\item Replace more than 3 repetitions of the same character (ie. \verb|grrrreat| becomes \verb|g xxrep r 4 eat|)
	\item Replace repetition at the word level (similar to above)
	\item Deal with ALL CAPS words replacing with a token and converting to lower case.
	\item Add spaces between special chars (ie. \verb|!!!| to \verb|! ! !|)
	\item Remove useless spaces (remove more than 2 spaces in sequence)	
	\item \textbf{Addition:} Move all text onto a single line by replacing new-lines inside a tweet with a reserved word (ie. \verb|\n| to \verb|xxnl|)
\end{enumerate} 
The following example shows the application of this data cleaning to a single tweet:
\begin{verbatim} 
Saber, entender y estar convencides que la frase \
#LaESILaDefendemosEntreTodes es nuestra linea es nuestro eje.\
#AlertaESI!!!!
Vamos por mas!!! e invitamos a todas aquellas personas que quieran \
se parte.
\end{verbatim}

\begin{verbatim} 
xxbos saber , entender y estar convencides que la frase \
# laesiladefendemosentretodes es nuestra linea es nuestro eje.\
xxnl  # alertaesi xxrep 4 ! xxnl vamos por mas ! ! ! e invitamos a \
todas aquellas personas que quieran se parte.
\end{verbatim}

\subsection{Tokenization}
We used sentencepiece~\cite{SentencePiece:DBLP:journals/corr/abs-1808-06226} to parse into sub-word units and reduce the possible out-of-vocabulary terms in the data set.  We selected a vocab size of 30,000 and used the byte-pair encoding (BPE) model. To our knowledge this is the first time that the BPE toenization has been used with ULMFiT in a competition model.

\section{Training and Results}
\label{sec:4}
\subsection{LM Training and Fine-tuning}
We train the LM using a 90/10 training/validation split, reporting the validation loss and accuracy of next-word prediction on the validation set. For the LM, we selected an ASGD Weight-Dropped Long Short Term Memory (AWD\_LSTM, described in Merity et al.\cite{Merity:DBLP:journals/corr/abs-1708-02182}) model included in Fast.ai. We replaced the typical Long Short Term Memory (LSTM) units with Quasi Recurrent Neural Network (QRNN, described in Bradbury et al.\cite{Bradbury:DBLP:journals/corr/BradburyMXS16}) units.  Our network has 2304 hidden-states, 3 layers and a softmax layer to predict the next-word.  We tied the embedding weights\cite{WeightTie:DBLP:journals/corr/PressW16} on the encoder and decoder for training.  We performed some simple tests with LSTM units and a Transformer Language model, finding all models were similar in performance during LM training. We thus chose to use QRNN units due to improved training speed compared to the alternatives. This model has about 60 million trainable parameters.  

Parameters used for training and fine-tuning are shown in Table \ref{tab:tab_training}.
For all networks we applied a dropout multiplier which scales the dropout used throughout the network.  We used the Adam optimizer with weight decay as indicated in the table.  

Following the work of Smith\cite{Smith:DBLP:journals/corr/abs-1803-09820} we found the largest learning-rate that we could apply and then ran a one-cycle policy for a single epoch.  This largest weight is shown in Table \ref{tab:tab_training} under "Learning Rate." Subsequent training epochs were run with one-cycle and lower learning rates indicated in Table \ref{tab:tab_training} under "Continued Training."

\begin{table}[ht]
	\caption{LM Training Parameters}
	\label{tab:tab_training}       
\begin{tabular}{lll}
	\hline\noalign{\smallskip}
	Param & LM & Fine-Tune LM \\
	\noalign{\smallskip}\hline\noalign{\smallskip}
	Weight Decay & 0.1 & 0.1 \\
	Dropout Mult & 1.0 & 1.0 \\
	Learning Rate & 1 epoch at $5*10^{-3}$ & 5 epochs at $3*10^{-3}$ \\
    Continued Training & 15 epochs at $1*10^{-3}$ & 10 epochs at $1*10^{-4}$\\
	\noalign{\smallskip}\hline
\end{tabular}
\end{table}

\subsection{Classification and Regression Fitting}
Again, following the play-book from Howard and Ruder\cite{HowardRuder:DBLP:journals/corr/abs-1801-06146}, we change the pre-trained network head to a softmax or linear output layer (as appropriate for the transfer task) and then load the LM weights for the layers below.  We train just the new head from random initialization, then unfreeze the entire network and train with differential learning rates.  We layout our training parameters in Table \ref{tab:clas_training}.

With the same learning rate and weight decay we apply a 5-fold cross-validation on the outputs and take the mean across the folds as our ensemble.  We sample 20 random seeds (see more in section \ref{sec:rand_seeds}) to find the best initialization for our gradient descent search.  From these samples, we select the best validation F1 metric or Mean Squared Error (MSE) for use in our test submission.
\subsubsection{Classifier setup}  For the classifier, we have a hidden layer and softmax head.  We over-sample the minority class to balance the outcomes for better training using Synthetic Minority Oversampling Technique (SMOTE, described in Chawla et al.\cite{Chawla:2002:SSM:1622407.1622416}).  Our loss is label smoothing as described in Pereyra et al.\cite{Labelsmoothing:DBLP:journals/corr/PereyraTCKH17} of the flattened cross-entropy loss.  In ULMFiT, gradual unfreezing allows us to avoid catastropic forgetting, focus each stage of training and preventing over-fitting of the parameters to the training cases.  We take an alternative approach to regularization and in our experiments found that we got similar results with label smoothing but without the separate steps and learning rate refinement required of gradual unfreezing.   
\subsubsection{Regression setup}  For the regression task, we fill all \verb|#N/A| labels with scores of 0.  We add a hidden layer and linear output head and MSE loss function. 

\begin{table}[ht]
	\caption{Classification and Regression Training Parameters}
	\label{tab:clas_training}       
	\begin{tabular}{ll}
		\hline\noalign{\smallskip}
		Param & Value \\
		\noalign{\smallskip}\hline\noalign{\smallskip}
		Weight Decay & 0.1  \\
		Dropout Mult &  0.7 \\
		Learning Rate (Head)& 2 epochs at $1*10^{-2}$\\
		Cont. Training & 15 epochs with diff lr:($1*10^{-3}/(2.6^4)$, $5*10^{-3}$)\\
		\noalign{\smallskip}\hline
	\end{tabular}
\end{table}

\subsection{Random Seed as a Hyperparamter}
\label{sec:rand_seeds}
For classification and regression, the random seed sets the initial random weights of the head layer. This initialization affects the final F1 metric achievable.  

\begin{figure}[ht]
	\includegraphics[width=0.75\textwidth]{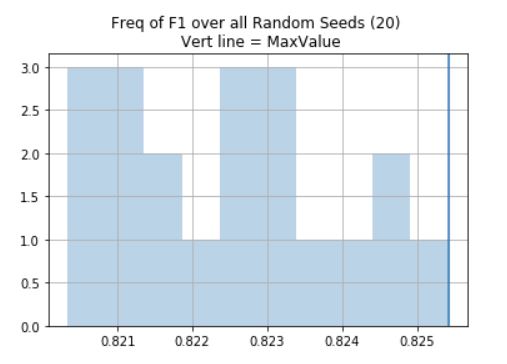}
	\caption{Histogram of F1 metric averaged across 5-fold metric}
	\label{fig:random_seed_hist}
\end{figure}

Across each of the 20 random seeds, we average the 5-folds and obtain a single F1 metric on the validation set. The histogram of 20-seed outcomes is shown in Figure \ref{fig:random_seed_hist} and covers a range  from 0.820 to 0.825 over the validation set. We selected our single best random seed for the test submission. With more exploration, a better seed could likely be found.  Though we only use a single seed for the LM training, one could do a similar search with random seeds for LM pre-training, and further select the best down-stream seed similar to Czapla et al \cite{Poleval:DBLP:journals/corr/abs-1810-10222}.

\subsection{Results}
Table \ref{tab:tab_results} gives three results from our submissions in the competition. The first is the baseline NBSVM solution, with an F1 of 0.7548.  Second is our first random seed selected for the classifier which produces a 0.8083 result.  While better than the NBSVM solution, we pick the best validation F1 from the 20 seeds we tried. This produced our final submission of 0.8099.  Our best model achieved an five-fold average F1 of 0.8254 on the validation set shown in Figure \ref{fig:random_seed_hist} but a test set F1 of 0.8099 - a drop of 0.0155 in F1 for the true out-of-sample data.  Also note that our third place entry was 1.1\% worse in F1 score than first place but 1.2\% \textit{better} in F1 than the $4^{th}$ place entry. 

\setlength{\tabcolsep}{8pt}
\begin{table}[ht]
	\caption{Comparative Results}
	\label{tab:tab_results}
	\begin{tabular}{lllll}
		\hline\noalign{\smallskip}
		 & Accuracy & Precision & Recall & F1 \\
		\noalign{\smallskip}\hline\noalign{\smallskip}
NBSVM &      0.8223 & 0.8180 & 0.7007 & 0.7548 \\
First Seed & 0.8461 & 0.7869 & 0.8309 & 0.8083 \\
Best Seed &  0.8458 & 0.7806 & 0.8416 & \textbf{0.8099} \\
		\noalign{\smallskip}\hline
	\end{tabular}
\end{table}

\section{Conclusion}
\label{sec:5}
This paper describes our implementation of a neural net model for classification and regression in the HAHA 2019 challenge.  Our solution placed $3^{rd}$ in Task 1 and $2^{nd}$ in Task 2 in the final competition standings.  We describe the data collection, pre-training, and final model building steps for this contest.  Twitter has slang and abbreviations that are unique to the short-format as well as generous use of emoticons.  To capture these features, we collected our own dataset based on Spanish Tweets that is 16 times larger than the competition data set and allowed us to pre-train a language model.  Humor is subtle and using a label smoothed loss prevented us from becoming overconfident in our predictions and train more quickly without the gradual unfreezing required by ULMFiT. We have open-sourced all  code used in this contest to further enable research on this task in the future.

\section{Author Contributions}BF was the primary researcher. PC contributed with suggestions for the random seeds as a hyper-parameters and label smoothing to speed up training. JH contributed with suggestion for higher dropout throughout the network for more generalization.

\section{Acknowledgements}
The author would like to thank all the participants on the fast.ai forums \footnote[9]{http://forums.fasta.ai} for their ideas and suggestions. Also, Kyle Kastner for his edits, suggestions and recommendations in writing up these results.

\bibliographystyle{splncs04}
\bibliography{local}   

\end{document}